\documentclass{article}

\usepackage{arxiv}

\usepackage[utf8]{inputenc} 
\usepackage[T1]{fontenc}    
\usepackage{hyperref}       
\usepackage{url}            
\usepackage{booktabs}       
\usepackage{amsfonts}       
\usepackage{nicefrac}       
\usepackage{microtype}      
\usepackage{lipsum}

\usepackage{graphicx}

\title{Federated Evaluation and Tuning for On-Device Personalization: System Design \& Applications}

\author{
  Matthias~Paulik\thanks{Corresponding author: <mpaulik@apple.com>}
  \And Matt~Seigel 
  \And Henry~Mason
  \And Dominic~Telaar
  \And Joris~Kluivers
  \And Rogier~van~Dalen
  \And Chi~Wai~Lau
  \And Luke~Carlson
  \And Filip~Granqvist
  \And Chris~Vandevelde
  \And Sudeep~Agarwal
  \And Julien~Freudiger
  \And Andrew~Byde
  \And Abhishek~Bhowmick
  \And Gaurav~Kapoor
  \And Si~Beaumont
  \And \'{A}ine~Cahill
  \And Dominic~Hughes
  \And Omid~Javidbakht
  \And Fei~Dong
  \And Rehan~Rishi
  \And Stanley~Hung
  \AND
  ~ \\ Apple
}

\begin{document}
\maketitle

\begin{abstract}
We describe the design of our federated task processing system. Originally, the system was created to support two specific federated tasks: evaluation and tuning of on-device ML systems, primarily for the purpose of personalizing these systems. In recent years, support for an additional federated task has been added: federated learning (FL) of deep neural networks. To our knowledge, only one other system has been described in literature that supports FL at scale. We include comparisons to that system to help discuss design decisions and attached trade-offs. Finally, we describe two specific large scale personalization use cases in detail to showcase the applicability of federated tuning to on-device personalization and to highlight application specific solutions.
\end{abstract}


\section{Introduction}
\label{intro}

Processing on end user devices (on-device processing) as opposed to server-based processing is a valued approach in enabling end user privacy. This strategy extends to many of our machine learned (ML) solutions, such as our predictive keyboard or on-device dictation. Personalization of these ML-based systems is often paramount in order to enable great user experiences. For example, the ability to personalize towards a user's vocabulary and language constructs is highly desirable in the context of both aforementioned systems.

Driven by these realities, we set out to address one specific on-device personalization use case around automatic speech recognition (ASR). This use case requires evaluating and tuning the global (i.e., common to all end users) parameters of a personalization algorithm that creates device specific ASR language models by ingesting data that is only available on-device. As explained in more detail in Section \ref{FET},  this initial use case ultimately led us to create a generic system that allows for evaluating and tuning ML systems across end user devices, i.e., a system that allows for `federated evaluation and tuning' (FE\&T). We generalized the system by abstracting away any use case specific bits, focusing on the requirements common to FE\&T, such as federated task distribution and execution, on-device evaluation data storage and task results ingestion and processing on a central server.

Federated learning (FL) refers to gradient based optimization of models across end user devices, with applied works of FL focusing on neural models \cite{DBLP:journals/corr/McMahanMRA16, DBLP:journals/corr/KonecnyMRR16, DBLP:journals/corr/KonecnyMYRSB16, DBLP:journals/corr/BonawitzIKMMPRS16, bonawitz2019federated, DBLP:journals/corr/abs-1908-07873, DBLP:journals/corr/abs-1912-04977}. The many FL publications of recent years influenced us to expand our system to better support FL. While we briefly describe these FL specific system additions, FL is not the primary subject of this paper. We instead focus on the application of our system to on-device personalization. Despite us not primarily focusing on FL, we believe FL and in particular the FL system design paper by \cite{bonawitz2019federated} to be the most relevant related work. We briefly discuss \cite{bonawitz2019federated} and other related work in Section \ref{related}. We continue to compare in more detail to \cite{bonawitz2019federated} in Section \ref{system} while describing our system, to help discuss design decisions and attached trade-offs. Section \ref{applications} describes two on-device ML system personalization use cases, including results obtained. We conclude with a brief summary in Section \ref{summary}.

\section{Related Work}
\label{related}

Even without its FL specific extensions, our system bears strong resemblance to the FL system described in \cite{bonawitz2019federated}. This similarity is due to the fact that both systems at their core require federated task distribution and execution, on-device evaluation/training data storage and server-side task results ingestion and processing. We will attempt to provide a more comprehensive comparison of both systems in Section \ref{system}, but want to point out one major difference already here. The system described in \cite{bonawitz2019federated} is custom designed for FL and does not allow for arbitrary distributed computation. The authors however mention in their future work section that they aim to generalize their FL system towards ``federated computation''. Our system was not motivated by and hence not custom built for training neural networks in a federated setting. Instead, on-device ML systems and their evaluation and tuning/personalization in a federated setting was the initial goal. Our system's on-device components therefore do not center around a neural model training library for task execution. Instead, implementation of on-device task execution is delegated to application specific plug-ins that communicate with our system's on-device task scheduling logic, data store and results reporting logic. The computation performed by these plug-ins can be arbitrary.

In addition to the aforementioned FL system paper, the larger field of FL \cite{DBLP:journals/corr/McMahanMRA16, DBLP:journals/corr/KonecnyMRR16, DBLP:journals/corr/KonecnyMYRSB16, DBLP:journals/corr/BonawitzIKMMPRS16, bonawitz2019federated, DBLP:journals/corr/abs-1908-07873, DBLP:journals/corr/abs-1912-04977} constitutes related work. This statement remains true even when only considering our system's FE\&T applications. Maybe trivially obvious, FL also requires model evaluation on held-out federated data. More interesting though is the notion of learning, and in particular what is being learned and how learning occurs in federated tuning (FT) compared to FL. FL learns the parameters of, at times large global neural models. In addition, assuming FL is based on federated averaging \cite{DBLP:journals/corr/McMahanMRA16}, learning (gradient descent) primarily occurs on-device, with the central server `simply' combining individual model updates. In FT, learning primarily occurs on the central server and is limited to a comparatively small set of global parameters (e.g., personalization algorithm parameters) that are `simply' evaluated across federated data. Our application example in Section \ref{sub_news} makes this concept more crisp. Finally, the fact that only small sets of global parameters are being learned in FT by centrally sharing evaluation metrics for specific parameter configurations minimizes the attack surface to end user privacy compared to FL, where high dimensional information (model updates) directly related to user data are centrally shared. We will touch on this topic again in Section \ref{system} when we sketch out our FL specific system extensions.

Given that the objective in FL is to learn a global neural network, only few FL publications address the topic of (model) personalization in a federated setting. \cite{DBLP:journals/corr/abs-1912-04977} however do include an insightful discussion on this specific topic. This discussion for example includes the observation that leveraging suitable user and context features fed into a global model can help to produce highly personalized model output. The discussion in \cite{DBLP:journals/corr/abs-1912-04977} also makes the connection to (federated) multi-task learning \cite{DBLP:journals/corr/SmithCST17, DBLP:journals/corr/abs-1906-06268}, when considering the local model learning problem as a separate task. In this context, \cite{DBLP:journals/corr/abs-1912-04977} observe that it also might be possible to apply techniques from multi-task learning where the task is to learn models over subsets of clients, which in turn could also yield more personalized model output. The authors of \cite{DBLP:journals/corr/abs-1912-04977} also discuss local model fine-tuning for the purpose of personalization and the connection to meta learning \cite{Thrun}, and cite in this context work by \cite{Jiang2019}, which shows that federated model averaging \cite{DBLP:journals/corr/McMahanMRA16} is equivalent to an algorithm called Reptile \cite{Nichol2018}. Reptile is a form of model-agnostic meta learning (MAML) \cite{Finn2017}. In general, in MAML, the idea is to initialize a model's parameters such that it can be adapted to a new domain, task or data as quickly as possible --- while making sure to avoid overfitting on the new task. Hence, models trained via federated averaging ought to be more amenable to being adapted/personalized subsequently on personal data. Finally, the authors of \cite{DBLP:journals/corr/abs-1912-04977} also point out prior work around fully decentralized personalization over end user devices, e.g., \cite{DBLP:journals/corr/VanhaesebrouckB16, DBLP:journals/corr/BelletGTT17, DBLP:journals/corr/abs-1803-09737}.

\section{FE\&T of ML Systems}
\label{FET}

\subsection{Motivation}

The conditions around our initial application (ASR personalization) led us to consider FE\&T of on-device ML systems and to ultimately implement our federated task processing system. The ASR system of our initial use case is a `conventional' ASR system, i.e., it uses multiple component ML models at inference time. In particular, the component ASR model subject to personalization is a statistical (non-neural) grammar that requires highly application specific code for modification. This explains our initial focus on ML systems versus neural models and also helps to explain our desire to provide flexibility in terms of on-device task execution implementation.

Further, in context of our initial use case, we operated under the assumption that only `sufficiently anonymized' and `siloed' ML data are available to us. Sufficiently anonymized in this context means that individual data points cannot, in a computationally reasonable manner, be tied back to a specific end user. Siloed ML data refers to the concept that ML data for a specific user cannot be aggregated across different, application specific data silos. These assumptions led us to focus on the questions of how to measure and how to improve user-specific ML system accuracy, especially in the context of personalization data that, in aggregate, remains inaccessible to us server-side. Our focus on on-device systems and their on-device evaluation and personalization (tuning) stems from these considerations. 

The consideration of federated evaluation (FE) and the attached server side aggregation of individual evaluation results was driven by the fact that on-device evaluation data is non-IID and often rather limited per device. Hence, individual evaluation results suffer from large quantities of uncertainty and only global results, aggregated across many devices offer meaningful insight. Finally, federated tuning presented itself as a natural and simple extension of federated evaluation. A crucial difference is that, instead of evaluating a single ML system, multiple ML systems are created on-device and evaluated, which might work as follows: A default ML system is set as baseline and a set of parameters sent from the server are applied to modify the behavior of the baseline ML system. Accuracy metrics for the baseline and the modified systems are sent back for analysis and to determine the next set of parameters to consider.

\subsection{The Challenge of Ground Truth}
\label{sub_truth}

In order to compute an evaluation metric on-device, we need to store or record an interaction with the to be evaluated ML system. The stored data needs to comprise the inputs to the ML system and the ground truth defining what the ML system's output should ideally have been. Recording the ground truth is often the more challenging aspect also because both positive and negative outcomes need to be covered. Trivially, such data can be collected at the same time the user interacts with the system, such that the tuple of inputs to the system and target output can be stored.

To illustrate this, we provide an example: App suggestions leverages the current device context (for example expressed as a vector of relevant variable values) to offer a list of suggested apps. Whenever a user opens an app, either from the home screen, via the search UI or via the app suggestion's UI, we can store on-device both, the current device context vector and the name of the app that was launched. Thus, we are obtaining both, the required app suggestions system input (the device context vector) as well as the name of the app that should ideally be offered prominently by the app suggestion UI.

Unfortunately, ground truth cannot always be obtained in such a simple manner. For example, in the context of evaluating ASR accuracy via the established word error rate metric, an error free reference transcript is required. Relying in this case on user interaction derived ground truth -- maybe whenever a user manually edits an automatic transcript -- is problematic, since evaluation data would be highly skewed towards such user caught and corrected error cases. One obvious solution might be to ask the end user to provide relevant ground truth, by for example asking the user to rate individual interactions. Although attractive from the perspective of potentially obtaining high-quality ground truth labels, such data cannot be collected automatically as in the case of the implicit signals, and relies heavily on the good will and participation of users. The quality of such labels is also highly dependent on the user, and may be inconsistent across the individuals providing such labels.

As illustrated by the ASR example, collecting ground truth labels on-device can be challenging and is sometimes even impossible. One potential remedy lies within leveraging a semi-supervised ML approach. For instance, one could conceive of a student-teacher training situation, where a slow, but highly accurate ML model is used to provide pseudo ground truth labels that are used to evaluate the actual run-time model. A related approach is to use a proxy evaluation metric that relies on ML system output confidences. In Section \ref{applications}, we will discuss in detail an ASR application specific solution of this form where ASR word confidences\footnote{Word confidences provide an estimate of how likely a word has been correctly recognized.} are used to compute a proxy evaluation metric. 

\section{System Description}
\label{system}

To help discuss design choices and associated trade-offs, we will in the following draw parallels to $Sys_{Bon}$, the FL system described in \cite{bonawitz2019federated}.

\subsection{High Level System Design}
\label{sub_highlevel}

\begin{figure} [h]
\vspace{3mm}
    \centering
    \includegraphics{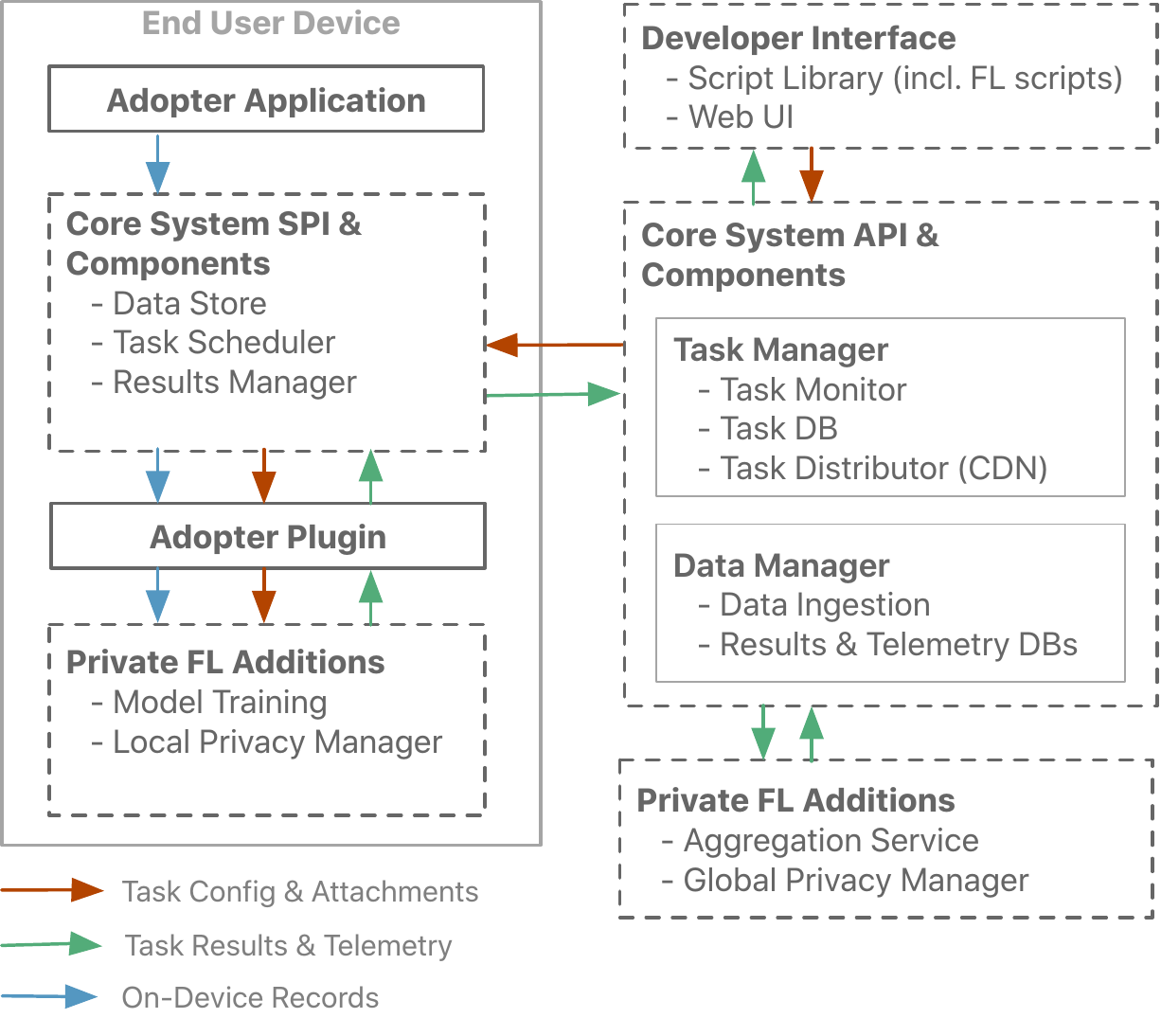}
    \caption{System Design Overview.}
    \label{fig:system}
\end{figure}

Figure \ref{fig:system} shows the high level system design. First to note, and in contrast to $Sys_{Bon}$ is the client side plug-in architecture. Task interpretation and execution is delegated to these application specific system plug-ins. The core system, described in more detail in Section \ref{sub_core} remains task agnostic. This overall system design therefore supports arbitrary federated tasks. Plug-ins that execute FL tasks are required to make use of the private federated learning (PFL) additions, which provide support for federated neural network training where final model updates undergo statistical noising that guarantee differential privacy (DP) \cite{DBLP:journals/fttcs/DworkR14}. More details on these PFL components are provide in Section \ref{sub_FL}.       

\subsection{Device Participation \& Data Handling}
\label{sub_part}

{\bf Device participation} is gated by both, system-level preconditions as well as federated task specific preconditions. The latter will be discussed in Section \ref{sub_core} in context of our task scheduler component. In terms of the former, participation is limited to genuine devices, guaranteed via a device attestation mechanism similar to $Sys_{Bon}$. In addition, devices need to be opted into the data collection program, thus enabling end users to opt out of participation. \cite{bonawitz2019federated} does not describe a similar mechanism. Our design choice to limit participation to opt-in devices has the potential to introduce bias. Another source for potential bias is the fact that devices need to be plugged into power, idle, and connected to an unmetered network in order to run federated tasks, which helps avoiding end users being negatively impacted by task execution and results reporting. $Sys_{Bon}$ sets the same preconditions and \cite{bonawitz2019federated} also discuss the potential for bias arising from these preconditions, while observing that mechanisms such as live A/B experiments of the final FL model can help detect any such bias. A/B experimentation (outside of our system's framework) is also a commonly employed by our adopter applications before fully deploying FL models or FT obtained personalization parameters. 

{\bf On-device data handling}. Our system provides an on-device data store that applications are recommended to use for logging user interaction data and associated ground truth. The data store is automatically purged on data collection program opt-out and write requests are denied if the device is opted out. To help protect against scenarios in which a user's device gets compromised, the data store remains encrypted until first user unlock after a device shutdown or reboot. In addition, the data store provides, per application configurable default implementations for common on-device data retention policies. The functionality provided includes sub-sampling (i.e., store only a random sample of user interactions), automatic deletion after $m$ days and capping the number of records by overwriting the oldest one once $n$ records have been written. These mechanisms help ensure that the amount of application specific data stored is limited and recent, and that no predictable sequence of events (e.g. the most recent ones) nor a long history of events are covered. In $Sys_{Bon}$, on-device data storage and attached retention policies are fully delegated to applications, which in turn have to ensure that their data is accessible for FL by implementing a $Sys_{Bon}$ specific API. The main motivator for us to offer a default on-device data store is ease of adoption. Since on-device computation on application specific data is delegated in our system to application specific plug-ins, there is no requirement for our (core) system to interpret or process this on-device data.

{\bf Results data handling}. While data held in our on-device data store never leaves the device, task results derived from this data, i.e., evaluation metrics (FE\&T) and statistically noised model updates (FL) are being send to our servers via encrypted channels. We also share information on the total amount of application specific data records held in the on-device data store as part of our system's health telemetry data, to help gauge the overall size of device populations available for application specific federated task execution. When logging data (results \& telemetry) server-side, we strip away all user and device identifies (e.g., IP addresses) to avoid that such server logged data can be tied back to a specific user or device. Data shared with our servers can be inspected by the end user within the privacy settings on-device. We will further touch on FL results data handling in Section \ref{sub_FL}.

\subsection{Core System}
\label{sub_core}

{\bf On-Device components} include \emph{data store} (see previous section), \emph{task scheduler} and \emph{results manager}. The \emph{results manager} is primarily responsible for sending back the task results and to populate an on-device DB that allows end users to inspect data shared with the server (see previous section for details). In addition to task results, the \emph{results manager} also collects and sends to the server system health related telemetry data.

The \emph{task scheduler} periodically connects to our content delivery network (CDN) to download a list of available task descriptors, if the system-level preconditions for device participation are met. Task descriptor download occurs in random order on a per registered plug-in basis to more evenly distribute the globally available compute. Once all task descriptors for a plug-in are downloaded, the scheduler samples exactly one matching (more details below) task for execution by the plug-in. The likelihood for task selection is governed by a sampling rate that is included in the task descriptor. While simple, this task selection mechanism provides an effective means to control how many devices globally attempt to execute a specific task. Task execution is further gated by our on-device matching logic, which examines additional preconditions listed in the task descriptor. These preconditions often refer to device-global variables, such as device or OS type and version. These preconditions can also refer to keys describing the on-device stored data. The latter allows, in a flexible manner, to run a specific task only on data falling out of specific end user interactions, e.g., speech audio recordings made in context of dictation vs. assistant requests. Once all task preconditions are met, the scheduler downloads the actual task including task specific attachments (e.g., models) and hands it over to the plug-in for execution. Logic is included to monitor task download and plug-in task execution times, as well as to abort task execution should a device for example become disconnected from power. 

Our described, on-device heavy task scheduling logic differs significantly from $Sys_{Bon}$, where task participation is orchestrated by the server. In $Sys_{Bon}$, devices periodically check in with the server and the server selects a subset of devices via reservoir sampling\footnote{\cite{bonawitz2019federated} mention that the selection protocol is amenable to more sophisticated methods.} to participate in a training round. Devices that are not selected receive instructions when to re-connect, thus regulating the pattern of device connections. This mechanism, referred to as `pace steering' by \cite{bonawitz2019federated} ensures that a sufficient number of devices connect to the server simultaneously, which, according to \cite{bonawitz2019federated}, is important to both, the rate of FL task progress and the properties of the Secure Aggregation \cite{DBLP:journals/corr/BonawitzIKMMPRS16} protocol which is optionally available in $Sys_{Bon}$. Compared to our simpler, task specific sample likelihood \footnote{Task sample likelihoods are configured server side.} approach, `pace steering' can therefore be viewed as more sophisticated but somewhat FL task and implementation specific optimization, focusing on synchronization amongst a fixed set of devices, connected simultaneously to the server. In our system design, we did not consider a task specific requirement of having to guarantee a minimum number of devices to connect simultaneously to the server. We primarily focused on having a sufficient number of devices reporting back, asynchronously, within a reasonable amount of time without overwhelming the server with too many concurrent connections. Any synchronization needs in our system are handled by a combination of a central results data base and coordinator scripts, as described in the following section.

{\bf Server components} include \emph{task manager}, \emph{data manager} and \emph{developer interface}. The \emph{task manager} is the core component of the server. It manages storage and publication of all tasks and their attachments (e.g., model) to the CDN as well as retiring tasks. Tasks are retired once either a task has been active for a pre-defined time window or once their targeted number of results have been received. Any straggling results arriving server side after a task has been retired are ignored. The \emph{task manager} also monitors the flow of incoming results to ensure that tasks that incur too much traffic do not overwhelm the infrastructure. Lastly, the component also ensures that results from older tasks are regularly and automatically purged from our central results database.

The \emph{data manager} drops any sensitive metadata from the incoming HTTP request and forwards results and telemetry information to a central data base. Telemetry data includes high level statistics around average task execution times, error conditions and amounts of available on-device data. The latter helps to gauge the size of federated device populations.

The \emph{developer interface} consists of both, a web UI and a python script library. The web UI is primarily used to monitor task status and to inspect telemetry data, but for example also allows to download results. System adopters primarily interact with the system via scripts to schedule tasks and to retrieve and process results available for each task. 

The server components of $Sys_{Bon}$ are not surprisingly highly specific and optimized to FL. Their implementation focuses on how to efficiently at scale do both, coordinating training rounds and performing aggregation for model averaging. Specifically, $Sys_{Bon}$ performs aggregation of results in memory. Only fully aggregated results are persisted. \cite{bonawitz2019federated} mention that ephemeral, in-memory aggregation improves latency but also minimizes the possibility of attacks within the data center, since no per-device result logs exist. Our design choice to persist per-device results and to delegate results aggregation to adopters stems from our task agnostic design, which requires flexibility. We consider computational inefficiencies and increases in attack surface stemming from this flexibility as less pressing in the context of FE\&T, given that only small dimensional information is centrally shared. The aforementioned considerations become however more pressing in the context of FL. We therefore pursued FL specific system additions to help address these considerations, as described in the next section.

\subsection{FL Specific Additions}
\label{sub_FL}

The many FL publications of recent years inspired us to extend our system to better support FL, using federated averaging \cite{DBLP:journals/corr/McMahanMRA16}. For FL applications, on-device task execution and server-side results processing is (on a high level) identical. Thus, we added common components that provide efficient implementations for both. For on-device task execution, a neural network training library provides \emph{model training} support. Server-side results processing is provided by a central \emph{aggregation service} that combines individual model updates to compute the averaged model of the current training iteration. Aggregation kicks off automatically as soon as a, by the \emph{FL training script} predefined number of individual model updates becomes available in the system's central data base. The \emph{FL training script} is responsible for coordinating the overall model training across multiple training iterations.

As discussed at the end of the previous section, storing individual model updates increases risk in the context of attacks in the data center. This is true since the individual model updates contain high dimensional information directly related to on-device stored data \cite{Fredrikson2015ModelIA, Melis2018InferenceAA}. We alleviate these risks by using both, a combination of ad hoc implementation strategies, as well as more formal strategies based on local and central (DP) \cite{DBLP:journals/fttcs/DworkR14}. In terms of a the former, individual model updates are encrypted on-device with a training round specific public key. The encrypted model updates are made available to the \emph{aggregation service} in random order. The \emph{aggregation service} holds the private key ephemerally and performs aggregation of the decrypted individual model updates in-memory. To add DP guarantees, the \emph{local privacy manager} adds a relaxed version of local DP noise to model updates \cite{bhowmick2019protection}. Server side, the \emph{global privacy manager} adds central DP noise to the final, updated model before making it available to the \emph{FL training script}, which coordinates training rounds. 

To summarize the differences to $Sys_{Bon}$, we use a central DB to temporarily store encrypted, local DP noised individual model updates. Further, we rely on potentially less efficient FL training scripts to coordinate training rounds. Finally, $Sys_{Bon}$ includes an implementation of secure aggregation, which helps to minimize the attack surface to end user privacy by allowing devices to share cryptographically masked updates. The cryptographic noise added to these individual model updates cancels out when computing the aggregate result. As pointed out by \cite{bonawitz2019federated}, secure aggregation however incurs costs that grow quadratically with the number of devices, effectively limiting the cohort sizes to hundreds of devices.


\section{On-Device Personalization Applications}
\label{applications}

Within our system, FL applications have gained traction in recent years. Examples include improving our acoustic keyword trigger models \cite{Granqvist20} or federated learning of  language models for an improved predictive keyboard \& error correction experience. However, applications around FE and FT still constitute a large percentage of system usage. Since FE occurs on historic user interaction records it often allows for significantly reduced turn-around times when compared to live A/B experimentation. For this reason, FE can help to quickly identify the most promising ML system or model candidates before exposing end users to these candidates via live A/B experimentation.

In the following, we describe two FT applications that center around ML system personalization, to highlight the applicability of FT to on-device personalization. The first application we describe is news personalization. Since on-device ground truth generation is less challenging in the context of news personalization (it can be approximated from user interactions with news content), we are focusing our description on the specific FT approach taken. This focus highlights how and where learning occurs in FT. It also highlights the flexibility our system provides in terms of on-device task execution implementation and server side, typically script based task results processing. The second application we are describing is ASR personalization, which was our initial use case, prompting us to develop our system in the first place. We consider the for this use case challenging problem of on-device ground truth generation as the more interesting aspect, and hence focus our description on this particular aspect.

\subsection{News Personalization}
\label{sub_news}

To help present news based on a user's interests, an on-device personalization algorithm is used. This algorithm is governed by several parameters, such as the half life of the time decay on an article's personalized score. Tuning of these parameters is challenging, since news content is constantly changing, e.g., due to new topics or due to seasonal variations in topic relevance. As a consequence, continuous adaptation of these parameters with quick turnaround times is important, so that the most relevant content can continue to be surfaced despite these changing trends. FT allows us to optimize these parameters quickly to achieve such turnaround times.

For news personalization, on-device evaluation/tuning data, including the notion of ground truth, can be derived from user interactions with news content. For example, we can store information on-device for articles a user has read (user tapped headline of news article; positive label), or has viewed but has not read (headline was not tapped; negative label). 

The on-device system plug-in for this use case accepts ranges of values for each parameter. For example, if the goal is to determine the optimal value of the half life parameter and the developer had no prior notion of what a good value should be, a range covering all possible valid values might be defined. During tuning task execution, the plug-in runs a randomized grid search on the parameter space defined, randomly generating configurations. The configurations are applied to the personalization algorithm which predicts how likely it is that a user will read an article contained in the on-device tuning data set. The predictions are compared the the ground truth labels to calculate a prediction loss for each randomly generated configuration.

Running a single FT task iteration results in millions of loss and configuration pairs from thousands of devices. These per iteration results are smoothed server-side by the application specific coordinator script in order to determine the search space for the next iteration, as described in the following. The problem of optimizing $n$ parameters can be implemented as an $n$-dimensional clustering problem with clusters of size $k$, where $k$ can be optimized. The cluster with the minimum prediction loss is identified by the following formula: 

\[loss = min(\sum_{\hat{r} \in R} \sum_{\hat{n} \in N} ||\hat{r} - \hat{n}|| \cdot loss(\hat{n}) \cdot \frac{1}{|N|})\]

$R$ is the set of clusters, and $N$ is the set of nodes within each clusters. This function finds the average loss of the cluster, weighted by the distance of each point to the centroid of the cluster, rewarding clusters that are tightly grouped together (and thus have lower variance) and have low prediction loss (and therefore have a higher accuracy in predictions). The cluster with the minimum prediction loss is used to determine the minimum and maximum values for each of the $n$ dimensions, and this forms the search space for the next FT task iteration. This process is repeated until convergence, yielding a locally optimal set of values for the parameters that were defined in the tuning task.

To prove the validity of this approach, we show results from two different FT runs. The FT runs differ in terms of the number of personalization algorithm parameters that are being optimized, as well as in how ground truth (positive and negative labels) was obtained in a rule based manner on-device (e.g., headline of article was tapped or not). Table \ref{table:dodEvals} shows these details, including results for the relative decrease in prediction loss. 

\begin{table} [h]
 \vspace{-2mm}
\caption{Federated tuning for news personalization.}
 \vspace{2mm}
\begin{center}
\begin{tabular}{ ccc } 
\bf{} &  \bf{FT Run 1} & \bf{FT Run 2} \\
\toprule
Iterations & 6 & 42\\
\cmidrule{1-3}
Parameters & 6 & 11\\
\cmidrule{1-3}
Pos. label & tapped & $>= n$ sec in article\\
\cmidrule{1-3}
Neg. label & not tapped & not tapped\\
\cmidrule{1-3}
Pred. loss & -86\% & -23\%\\
\bottomrule
\end{tabular}
\end{center}
\label{table:dodEvals}
\end{table}

For both FT runs, live A/B experiments were conducted to measure the impact of the optimized personalization algorithm parameters on end user experience. Table \ref{table:AB} shows the results. 

\begin{table} [h]
 \vspace{-2mm}
\caption{Live A/B experimentation results.}
 \vspace{2mm}
\begin{center}
\begin{tabular}{ ccc }
&  \multicolumn{2}{c}{\bf{Delta[\%]}} \\
\cline{2-3}
\bf{} &  \bf{Run 1} & \bf{Run 2} \\
\toprule
daily article views & +1.98 & +1.87\\
\cmidrule{1-3}
daily time spent & n/a & +0.90\\
\bottomrule
\end{tabular}
\end{center}
\label{table:AB}
\end{table}

The optimized parameters from FT run 1 resulted in a 1.98\% increase in daily article views, but no statistically significant difference in daily time spent within the application\footnote{Stocks application, which includes news headlines.}. The optimized parameters from FT run 2 resulted in a 1.87\% increase in the daily article views, and a 0.90\% increase in the daily time spent within the application.

\subsection{ASR Personalization}
\label{sub_pasr}

In the following, we are describing a `hybrid' ASR system solution, distributed across end user device and server. This hybrid ASR system computes its final recognition result using server-side ASR system combination, in which the recognition result from a highly personalized on-device ASR system is combined with the results from a more generic server based ASR system. Similar hybrid approaches to ASR personalization have been explored before \cite{georges2014,mcgraw2016,glackin2017}, albeit not to our knowledge in the context of at scale FT of the personalized on-device ASR system.

Our ASR engine that transcribes a user's speech input utilizes a generic vocabulary consisting of several hundred thousand words. In order to enable recognition of out-of-vocabulary words and less common language constructs a specific user might utter, the ASR system is able to dynamically load user-specific statistical grammars at run-time using the approach described in \cite{paulik2016method}, which is similar to \cite{novak-2012-look} and the more general class-based language modeling concept described in \cite{brown-92-class}. These dynamic grammars are spliced into the larger, generic (i.e. non-user specific) grammar of the on-device ASR system. The user-specific grammars can be learned on-device from user specific text data sources. These on-device data sources include for example a user's contact list and contact interaction counts, internet search history, frequently visited addresses/points of interest, vocabulary learned from keyboard interactions, named entities discovered in read web pages and news articles, etc. The overall personalization algorithm leverages around 30 parameters that govern both, user-specific grammar compilation and their dynamic application at run-time. For example, these parameters include per sub-grammar global scaling factors, determine what on-device text data sources to use (e.g. based on usage) and influence word n-gram probability estimation from frequency counts (e.g contact call frequencies).

FT relies on FE, which in turn requires evaluation data that includes ground truth labels. The raw evaluation data falls out of regular user interactions with the ASR system, for example when a user addresses our assistant or dictates a text message. We sample some of these interactions, safely storing the speech audio in our on-device data store. Ground truth generation is however more complicated. The established evaluation metric in ASR is word error rate (WER) and our objective is to reduce WER. Word error rate is defined as the ratio of the word level Levenshtein distance \cite{Levenshtein} to number of words in the reference transcription:


\begin{center}
$WER = \frac{\# minimum\ word\ edits}{\# reference\ words}$
\end{center}

That is, to compute WER we require our evaluation data to be labeled with manual reference transcriptions, which are not readily available on-device (see also the discussion in Section \ref{sub_truth}). We therefore decided to rely on a semi-supervised learning approach that leverages a machine learned, global word confidence model \cite{evermann-2000-conf,wessel2001,sanchis2003,jiang2005,seigel2011} to estimate word error rate (eWER) in the absence of reference transcriptions.

Word confidence models assign noisy probability scores $c(W, D) \in [0,1]$ to each hypothesized word $W$, expressing the likelihood for the word being correctly recognized. To do so, these models use input features $D$ that are typically computed during the ASR decoding process. In the context of this work, it is important to only rely on features $D$ that remain largely independent from personalized grammar influences, e.g. acoustic likelihood scores and phone/word duration features. We estimate the WER for an utterance $\mathbf{U}$ of length $|\mathbf{U}|$ as follows:

\begin{center}
$eWER(\mathbf{U}) = 100 \times \left(1 - \frac{1}{|\mathbf{U}|}\sum_{W \in \mathbf{U}}{c(W, D)}\right) + \rho$
\end{center}


The global calibration factor $\rho$ compensates for the fact that our confidence model cannot account for ASR word deletions.

Table \ref{table:ewer} shows how well our eWER metric compares with actual WER measurements. The results shown are in context of a simplified form of personalization towards a single data source, namely contact names in a user's address book without frequency counts. Comparison of estimated and actual WER is possible for this specific type of personalization due to the existence of a small server-side test set that includes contact names derived personalization grammars on a per utterance basis. The first data row in Table \ref{table:ewer} shows results on this centrally held test. Since human created reference transcriptions are available for this test set, we can directly compare actual WER with eWER. The table also lists results for three differently sized, federated test sets. Reference transcriptions, and therefore actual WER values are not available for these federated sets. In fact, these federated test sets remain hidden to us, i.e. we have no access to the underlying audio data nor the associated transcript. If we assume that our centrally held test set is drawn from the same distribution as the federated test sets, WER and eWER values on all these sets should be similar.

\begin{table} [h]
 \vspace{-2mm}
\caption{Comparing estimated and actual WER for a simplified form of contact list personalization.}
 \vspace{2mm}
\begin{center}
\begin{tabular}{ cccc } 
 \multicolumn{2}{c}{} & \multicolumn{2}{c}{\bf{WER [\%]}} \\
\cline{3-4}
\bf{evaluation} & \bf{\# requests} & \bf{actual} & \bf{estimated} \\
\midrule
central & 3,654 & 12.0 & 12.1\\ 
\cmidrule{1-4}
& 10k & N/A & 14.4\\
\cmidrule{2-4}
federated & 100k & N/A & 13.2 \\
\cmidrule{2-4}
& 300k & N/A & 13.1 \\ 

\end{tabular}
\end{center}
\label{table:ewer}
\end{table}

The results on the differently sized federated tests sets show that eWER stabilizes at around 13\% at larger set sizes, comprising 100k and more utterances. We confirmed this behavior via multiple federated evaluation runs (not shown in the table). The need for such rather large federated test set sizes is largely due to the noise in the word error estimates of the machine learned confidence model. However, even at federated test set sizes of 100k and beyond, we still observe a difference of 1\% absolute between the eWER of the federated set and the eWER as well the actual WER observed on the central, human transcribed test set. We determined that this difference stems from the fact that our central test set had been scrubbed of audio recordings that only contain background noise and no audible speech. 

Based on the objective function of minimizing eWER, we tune via FT the global parameters that govern how each device personalizes its client ASR system from personal data sources. 
As mentioned at the beginning of this section, we use server-side ASR system combination to combine the result of the on-device ASR system with the result of the server-side ASR system. Thus, the final recognition result is available to us server side, enabling us to measure the impact of the optimized personalization algorithm parameters on the final transcription accuracy in terms of actual WER. To do so, we periodically and randomly sample, from production traffic, several thousand anonymized speech recognition results after ASR system combination and ask human graders to provide reference transcriptions after listening to the temporarily available speech audio. ASR system combination only affects the outcome of approximately 10\% of speech utterances directed to our assistant. Overall, we observe that mostly those utterances are affected by ASR system combination for which the generic server ASR system struggles, i.e. where transcription accuracy suffers from a significantly elevated WER compared to the remaining $\sim$90\% of the speech recognition traffic.

The results shown in Table \ref{table:finalASR} relate to 3,851 assistant directed requests where the final automatic transcript changed due to ASR system combination. We can automatically detect utterances for which the final transcript includes words or phrases surfaced by our on-device personalized grammar components. In the following, we refer to these utterances as `user vocabulary' utterances. Utterances for which the final result changed due to ASR system combination but the transcript only includes words and phrases that stem from our generic grammars are in the following referred to as `generic vocabulary' utterances.  The first data column in Table \ref{table:finalASR} shows the request count, and the second data row shows the WER that would have been achieved without any personalized on-device ASR. The third data row shows the WER after ASR system combination with the client ASR result. Taking overall traffic into account, roughly 3.9\% (6.1\%) of all assistant directed utterances are user (generic) vocabulary utterances benefiting from personalized client ASR. Especially on user vocabulary utterances, the relative WER reductions of 16.4\% are significant. However, even for generic vocabulary utterances, small relative WER reductions of 1.4\% are achieved, thanks to generic ASR system combination effects.

\begin{table} [h]
 \vspace{-2mm}
\caption{Word error rates on utterances affected by system combination.}
 \vspace{2mm}
\begin{center}

\begin{tabular}{ lcccc } 
\toprule
& &  \multicolumn{2}{c}{\bf{WER[\%]}} \\
\cline{3-4}
\bf{Requests with} & \bf{Count} & \bf{Server ASR} & \bf{+Client ASR} \\
\midrule
user vocab. & 1,504 & 24.4 & 20.1 \\
generic vocab. & 2,347 & 14.6 & 14.4 \\
\end{tabular}
\end{center}
\label{table:finalASR}
\end{table}

\section{Summary}
\label{summary}

We described the system design of our federated task processing system, which has its roots in federated evaluation (FE) and federated tuning (FT) of ML systems. System development was originally motivated by a specific use case: On-device personalization of an automatic speech recognition (ASR) system, which requires optimization of a global parameter vector that governs  how the ASR system is personalized on an end user's device. We provided a detailed description of this initial use case and of one additional personalization use case, highlighting the applicability of FT to on-device ML system personalization. Largely due to the need to support arbitrary evaluation task and tuning task implementations in context of ML systems, our federated task processing system design allows in principle for arbitrary distributed computation. The ability to support arbitrary distributed computation is is partly demonstrated by the described personalization applications, as well as the federated learning (FL) specific systems extensions we have sketched out. These system extensions provide FL support combined with DP guarantees and are thus referred to as `private federated learning' (PFL) extensions. 

\bibliographystyle{unsrt}  
\bibliography{references}  

\end{document}